\documentclass[runningheads]{llncs}
\pdfoutput=1
\usepackage[T1]{fontenc}
\usepackage{graphicx}
\usepackage{amsfonts}
\usepackage{amsmath}
\usepackage{xcolor}
\usepackage{marvosym}
\usepackage{orcidlink}
\begin{document}

\title{Large Scale Unsupervised Brain MRI Image Registration Solution for Learn2Reg 2024}
%
%
\author{
	Yuxi Zhang\inst{1,2}\orcidlink{0009-0000-7190-3467} \and 
	  Xiang Chen\inst{1,2}\textsuperscript{(\Letter)}\orcidlink{0000-0003-4203-4578}\and 
        Jiazheng Wang\inst{1,2}\orcidlink{0000-0003-2534-4232} \and 
	Min Liu\inst{1,2}\orcidlink{0000-0001-6406-4896}	\and 
	Yaonan Wang\inst{1,2}\orcidlink{0000-0002-0519-6458
} \and
        Dongdong Liu\inst{3}\orcidlink{0000-0001-5265-6907} \and
        Renjiu Hu\inst{4}\orcidlink{0000-0003-4651-3349} \and
        Hang Zhang\inst{4}\orcidlink{0000-0003-0115-387X} 
        }

\authorrunning{Y. Zhang et al.}
%
\institute{College of Electrical and Information Engineering, Hunan University, Changsha, Hunan, China \and
National Engineering Research Center of Robot Visual Perception and Control Technology, Hunan University, Changsha, Hunan, China
\email{\{hnuzyx, xiangc, wjiazheng, liu\_min, yaonan\}@hnu.edu.cn}
\and New York University, New York, USA. \email{ddliu@nyu.edu}
\and Cornell University, New York, USA. \email{\{rh656,hz459\}@cornell.edu}}
\maketitle   
%
\begin{abstract}
In this paper, we summarize the methods and experimental results we proposed for Task 2 in the learn2reg 2024 Challenge. This task focuses on unsupervised registration of anatomical structures in brain MRI images between different patients. The difficulty lies in: (1) without segmentation labels, and (2) a large amount of data. To address these challenges, we built an efficient backbone network and explored several schemes to further enhance registration accuracy. Under the guidance of the NCC loss function and smoothness regularization loss function, we obtained a smooth and reasonable deformation field. According to the leaderboard, our method achieved a Dice coefficient of 77.34\%, which is 1.4\% higher than the TransMorph. Overall, we won second place on the leaderboard for Task 2.
\end{abstract}
\section{Introduction}
Image registration stands as an indispensable procedure within the realm of medical image analysis, fundamentally tasked with aligning a set of images to a common reference frame~\cite{chen2021deepSurvey}. Particularly in the context of brain imaging, this technique underpins a multitude of critical applications, including image segmentation, surgical planning with image guidance, and the diagnosis of various diseases. Traditional approaches to registration, while effective, are often marred by their laborious nature, necessitating numerous iterative steps that consume considerable time.

In recent years, the advent of deep learning has ushered in a new era of medical image registration methods. These novel approaches, as proposed in works such as \cite{chen2022transmorph,tian2024unigradicon,hoffmann2022synthmorph,xu2016evaluation,balakrishnan2019voxelmorph,chen2021deepDiscontinuity,zhang2023spatially}, promise not only expedited registration times but also performance that rivals or surpasses that of conventional techniques. However, as the sophistication of deep learning models increases, so too does the complexity and the number of parameters involved, potentially complicating the registration process.

The Learn2Reg 2024 MICCAI Grand Challenge, specifically Task 2 LUMIR, presents an opportunity to evaluate the scalability and efficiency of registration methods through a large-scale dataset designed for inter-subject registration. 

In this paper, we delve into the contributions of these architectural innovations to the accuracy and efficiency of a registration backbone, aiming to elucidate their role in enhancing the performance of medical image registration in a large-scale dataset. 
\section{Methodology}
\subsection{Proposed Method}
\begin{figure}
    \centering
    \includegraphics[width=1\linewidth]{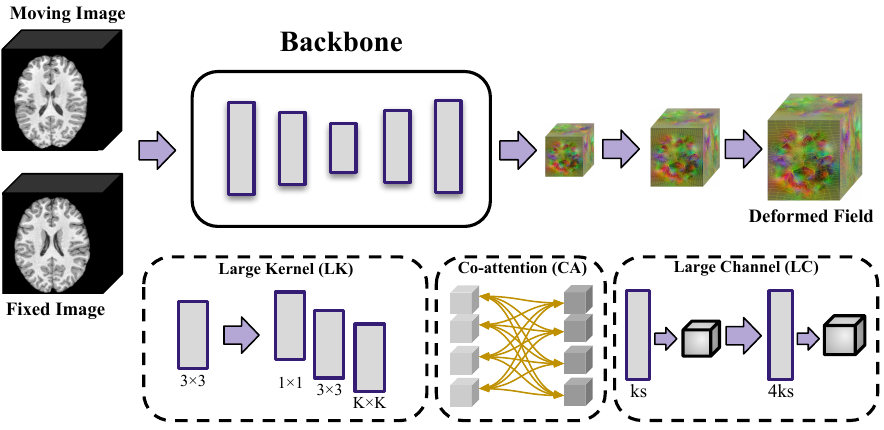}
    \caption{The backbone network includes a general convolutional block and cascade registration structure. To enhance its efficiency, several schemes/structures are incorporated as shown at the bottom of the figure.}
    \label{fig: network}
\end{figure}

To ensure efficient registration performance on large-scale brain MR datasets, we have constructed a robust backbone network, inspired by the works of Zhang et al.~\cite{zhang2024memwarp,zhang2024slicer}, as depicted in Figure.\ref{fig: network}. Our network architecture comprises an encoder with a series of convolutional layers and a cascaded mechanism designed to refine the prediction of deformation fields.

In pursuit of enhanced registration accuracy, we have integrated several innovative schemes and architectural elements into our backbone. These include:

\textbf{Co-Attention Mechanism (CA)}
We implement a co-attention mechanism to facilitate feature integration at the lower layers of the encoder, drawing from the work of Chen et al.~\cite{chen2021car}.

\textbf{Large Kernel Convolution (LK)}
We replace conventional convolutions in the backbone with large kernel convolutions, as proposed by Jia et al.~\cite{jia2022u}, to capture broader contextual information.

\textbf{Bilateral Filtering (BF)}
Post-deformation field prediction, we apply an image-guided bilateral filtering technique, following the approach of Wagner et al.~\cite{wagner2022trainable}, to refine the registration results.

\textbf{Large Convolution Channels (LC)}
We expand the number of convolutional channels in the deformation field prediction stage, building upon the findings of Zhang et al.~\cite{zhang2024memwarp,zhang2023spatially}, to enhance the generation of the deformation field.

Each of these elements is meticulously selected and integrated to bolster the backbone's capacity for accurate and efficient registration. The synergistic effect of these enhancements is expected to yield a significant improvement in performance.

\subsection{Loss Function}
\subsubsection{Total Loss.} The total loss function($\mathcal{L}$) basically consists of similarity loss($\mathcal{L}_{NCC}$) and regularization loss($\mathcal{L}_{Reg}$):
\begin{equation}
\mathcal{L} =\lambda _0\mathcal{L} _{NCC}+\lambda _1\mathcal{L} _{Reg}.
\end{equation}

\subsubsection{NCC Loss.} 
For the unsupervised training of our network, we have chosen the Normalization Local Correlation Coefficient ($\mathcal{L}_{NCC}$) as our similarity loss function. This metric is preferred for its robustness to intensity variations and sensitivity to image structure, which are crucial for medical image registration. Compared to the Mean Square Error, our experiments demonstrate that $\mathcal{L}_{NCC}$ significantly improves registration accuracy.

\begin{equation}
NCC\left( x,y \right) =\frac{\sum_{i=1}^n{\left( x_i-\overline{x} \right) \left( y_i-\overline{y} \right)}}{\sqrt{\sum_{i=1}^n{\left( x_i-\overline{x} \right) ^2}}\sqrt{\sum_{i=1}^n{\left( y_i-\overline{y} \right) ^2}}},
\end{equation}
where the $\overline{x}$ and $\overline{y}$ represent the mean of the sample $x,y$. 
During training, the negative normalized cross-correlation (NCC) is used to minimize the loss function $\mathcal{L}_{NCC}$, where a higher NCC value indicates greater similarity between images.

\subsubsection{Regularization Loss.} 
The regularization loss($\mathcal{L}_{Reg}$) which is based on its spatial gradient ensures the smoothness of the deformation field and avoids unrealistic deformations:
\begin{equation}
\mathcal{L} _{Reg}=\left\| \nabla \boldsymbol{u} \right\| _{2}^{2}.
\end{equation}

\section{Experimental Setup and Results}
\subsection{Datasets}
For Task 2, our research employed the OpenBHB dataset~\cite{dufumier2022openbhb}, which comprises T1-weighted brain MRI scans meticulously curated from a consortium of 10 public datasets~\cite{taha2023magnetic}. Notably, a subset of this comprehensive collection originates from the AFIDs project, leveraging the foundational OASIS dataset~\cite{marcus2007open}. The dataset encompasses 3384 meticulously prepared training samples and a robust set of 38 validation samples. To standardize the data, all images have been meticulously converted to NIfTI format, resampled, and precisely cropped to the region of interest, aligning to uniform dimensions of $160 \times 224 \times 192$ pixels with a consistent voxel spacing of 1 mm isotropic resolution. Task 2 is distinctive for its emphasis on unsupervised image registration; as such, segmentation labels and landmarks for the training data are intentionally withheld. To aid participants, the organizers have graciously provided a curated subset facilitating preliminary sanity checks, available via their dedicated GitHub repository\footnote{\url{https://github.com/JHU-MedImage-Reg/LUMIR\_L2R}}.

\subsection{Experimental Setup}
Our network is trained in an unsupervised manner, utilizing a dataset comprising 3374 unlabeled samples for training and a separate validation set of 10 labeled samples reserved exclusively for model evaluation and checkpointing.

The training is conducted on an Nvidia A6000 server, leveraging the PyTorch framework with a fixed epoch size of 200. We employ the Adam optimizer, initiating the learning rate at $4e-4$ with a batch size of one. For the loss function, the NCC coefficient $\lambda_0$ is assigned as 1, while the regularization term's coefficient $\lambda_1$ is assigned as 6, ensuring a balanced optimization process.

\subsection{Results}
The Learn2Reg 2024 Grand Challenge offers a comprehensive set of quantitative indicators to rigorously assess the performance of registration algorithms. For Task 2, in order to conduct a comprehensive evaluation, it calculates the Dice coefficient, target registration error (TRE), non-diffeomorphic volumes (NDV)\cite{liu2024finite} and 95\%Hausdorff distance(HD95).

The results of our quantitative assessment on the validation set are meticulously detailed in Table~\ref{tab1}. Notably, the metric `Zero Displacement' indicates the absence of deformation, serving as a benchmark for our analysis. Our innovative approach demonstrates a marked enhancement over the state-of-the-art baseline method, TransMorph. Specifically, we report a 1.4\% increase in the Dice coefficient, signifying a superior overlap of the segmented regions. Furthermore, our method yields a 0.3508 reduction in the NDV index, underscoring a substantial decrease in non-diffeomorphic volumes and thus preserving the integrity of anatomical structures. Additionally, a 0.1659 improvement in the HD95 metric reflects a notable reduction in the maximum surface distance, indicative of heightened registration precision.

Our findings, presented in Table~\ref{tab2}, reveal the impact of integrating diverse schemes into the backbone network. The incorporation of BF post-registration has led to a marked improvement in the Dice and HD95 metrics, enhancing deformation smoothness. Our backbone network's augmentation with LK Convolution modules and an increased channel count has bolstered TRE and NDV metrics. Incorporating a CA module into the backbone network, complemented by bilateral filtering, enhances the Dice performance on both the validation and leaderboard. While LC has shown larger improvements in the validation set, their impact on the leader-board performance was not significantly different, indicating a need for further exploration to ensure consistent enhancement across all sets. The difference between the backbone+LC-1 and backbone+LC-2 lies in the deformation field cascade scheme. A better cascade registration scheme can lead to better registration accuracy and smoother deformation fields, as evidenced by the results of backbone+LC-2.

\begin{table}
\caption{Results of our method on the validation set in the Learn2Reg 2024 MICCAI Grand Challenge. More validation results could be found on the challenge leaderboard. The best results are highlighted in bold.}\label{tab1}
\centering
\begin{tabular}{c|c|c|c|c}
\hline
Method &  Dice$\uparrow$ & TRE(mm)$\downarrow$ & NDV(\%)$\downarrow$ & HD95(mm)$\downarrow$\\
\hline
Zero Displacement(Before Reg) & $56.57\pm 2.63$ & 4.3543 & 0.0000 & 4.7876\\
TransMorph\cite{chen2022transmorph} &  $75.94\pm 3.19$ & 2.4225 & 0.3509 & 3.5074\\
uniGradICON\cite{tian2024unigradicon} &  $73.69\pm 4.12$ & 2.5727 & 0.0000 & 3.6107\\
SynthMorph\cite{hoffmann2022synthmorph} & $72.43\pm 2.94$ & 2.6099 & 0.0000 & 3.5730\\
VoxelMorph\cite{balakrishnan2019voxelmorph} & $71.86\pm 3.40$ & 3.1545 & 1.1836 & 3.9821\\
deedsBCV\cite{xu2016evaluation} & $69.77\pm 2.74$ & \textbf{2.2230} & 0.0001 & 3.9540\\
\textbf{Ours(next-gen-nn)} & $\mathbf{77.34\pm 3.13}$ & 2.3360 & 0.0001 & \textbf{3.3415}\\
\hline
\end{tabular}
\end{table}

\begin{table}
\caption{The results of incorporating different schemes into backbone network. Note that, 'sc' denotes the number of channels in the convolutions of the backbone network. In general, larger channels would lead to higher registration accuracy. The best results are highlighted in bold.}\label{tab2}
\centering
\begin{tabular}{c|c|c|c|c|c}
\hline
Method & Val Dice & Dice$\uparrow$ & TRE(mm)$\downarrow$ & NDV(\%)$\downarrow$ & HD95(mm)$\downarrow$\\
\hline
Before Reg & $56.57$ & $56.57\pm 2.63$ & 4.3543 & 0.0000 & 4.7876\\
backbone(sc:32) &  76.50 & $77.31\pm 3.11$ & 2.3753 & 0.2177 & 3.3437 \\
backbone+BF(sc:32) & 76.53 & $\mathbf{77.34\pm 3.12}$ & 2.3734 & 0.1861 & \textbf{3.3382}\\
backbone+CA(sc:32) & 76.52 & - & - & - & -\\
backbone+CA+BF(sc:32) & 76.54 & $77.33\pm 3.03$ & 2.3833 & 0.1845 & 3.3646\\
backbone+LK(sc:36) & 76.58 & $77.25\pm 3.11$ & 2.3527 & 0.1530 & 3.3636\\
backbone+LC-1(sc:64) & 76.66 & $77.33\pm 3.11$ & 2.3852 & 0.1854 & 3.3542\\
backbone+LC-2(sc:64) & \textbf{76.76} & $\mathbf{77.34\pm 3.13}$ & \textbf{2.3360} & \textbf{0.0001} & 3.3415\\
\hline
\end{tabular}
\end{table}

\begin{figure}[!h]
    \centering
    \includegraphics[width=1\linewidth]{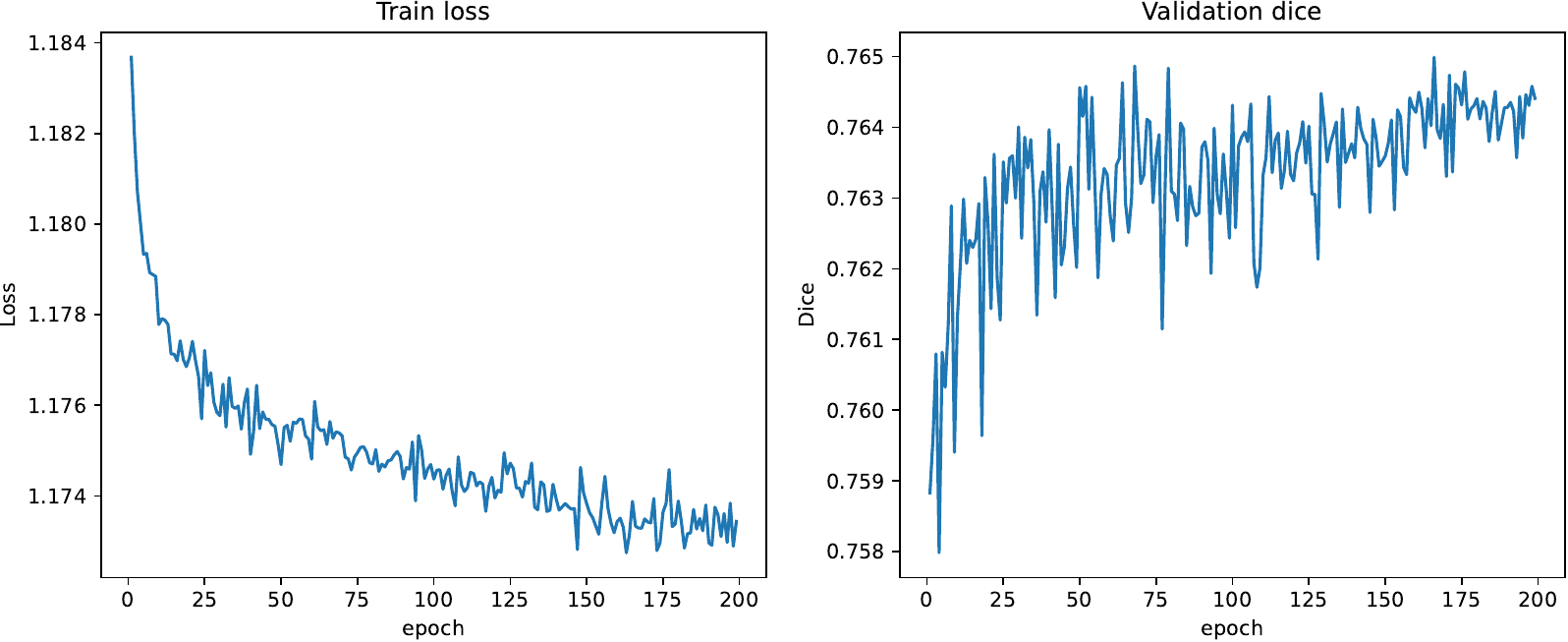}
    \caption{The loss curve and dice coefficients on validation set during the training process}
    \label{fig:loss figure}
\end{figure}

Fig.\ref{fig:loss figure} illustrates the progression of our model's training loss (left panel) and its validation performance (right panel) across successive epochs. The data clearly demonstrate that our approach achieves rapid convergence and maintains robust performance on the validation dataset, only after a minimal number of initial epochs.

\begin{figure}[!h]
    \centering
    \includegraphics[width=1\linewidth]{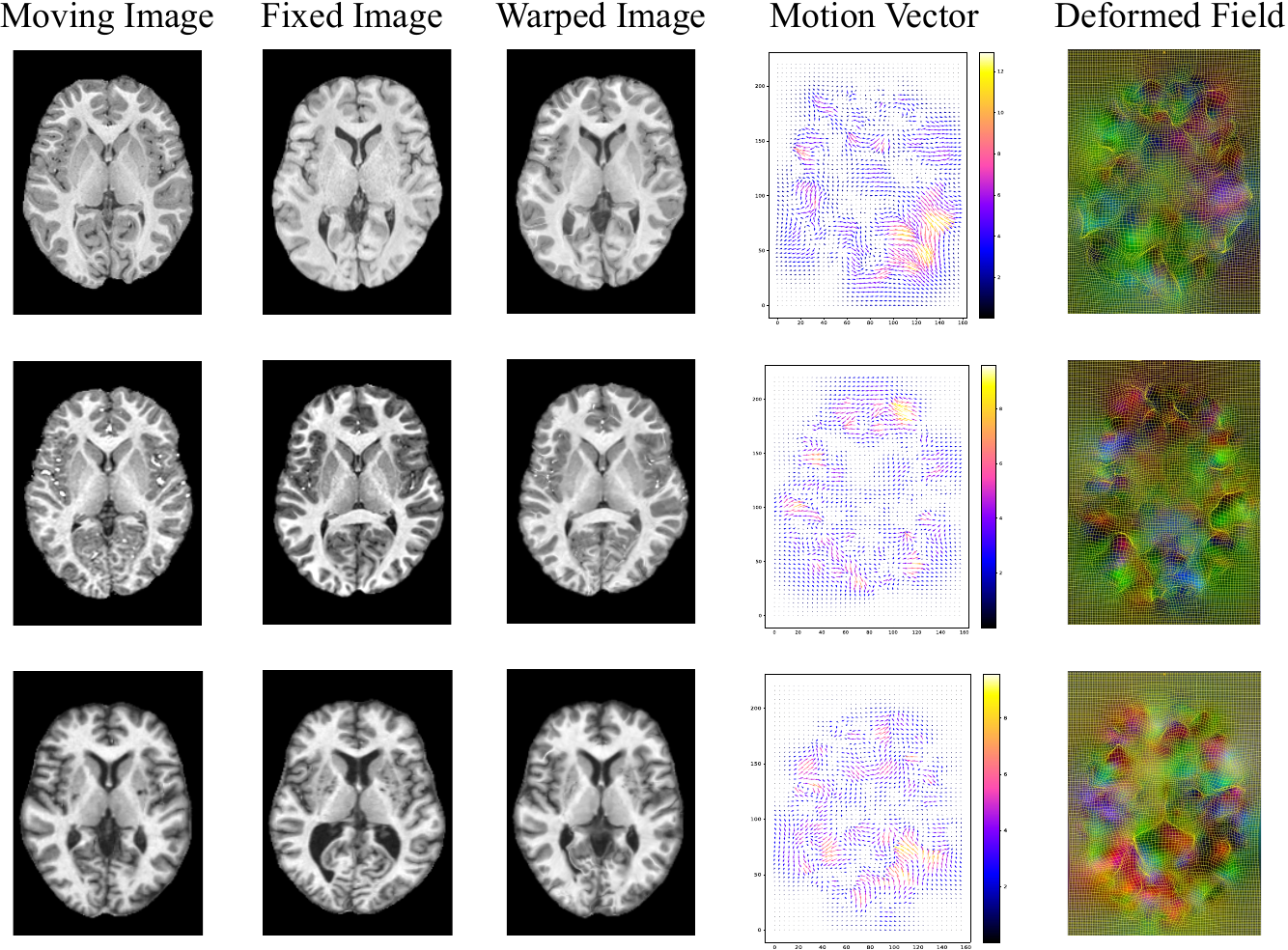}
    \caption{Visualization of three test samples in the submitted data. From left to right are the moving image, fixed image, predicted motion vector, and predicted deformation field.}
    \label{fig:vector visualization}
\end{figure}

Fig.\ref{fig:vector visualization} presents the visualization of deformation fields for three representative pairs among the 38 validation samples. The deformation is depicted through two distinct methods: streamlines representing motion vectors and grids illustrating the extent of deformation. The sequence of images, from left to right, displays the moving image, the corresponding fixed image, the warped moving image, the motion vectors, and the deformation grid.

\section{Analysis and Conclusion}
After thoroughly analysing the registration results, we observe several notable challenges in the LUMIR dataset:
(1) An imbalanced distribution across training, validation, and test sets is evident. This discrepancy is underscored by the observation that some methods perform admirably on the validation set but exhibit no significant improvement or even a decline in performance on the public leaderboard.
(2) There is a significant variance in registration difficulty among different pairs of registrations. The leaderboard results demonstrate a wide range of registration Dice scores, spanning from 18\% to 99\%. To enhance registration efficiency further, a deeper investigation into the particularly challenging cases, such as those with a Dice score of 18\%, would be beneficial. However, such an exploration lies beyond the scope of the present paper.

In this study, we demonstrate that in large-scale data scenarios, the efficiency of a backbone network is pivotal to registration performance. Other schemes or architectures, despite their proven utility in prior research, merely achieve comparable results to the backbone, as evidenced by the leaderboard standings.

\begin{credits}
\subsubsection{\ackname} 
    Thanks all the organizers of the MICCAI 2024 Learn2Reg challenge. This work was supported in part by the National Key Research and Development Program of China (grant number 2022YFE0134700), and in part by the National Natural Science Foundation of China (grant number 62221002), and in part by the Fundamental Research Funds for the Central Universities, China.
\end{credits}

%
%
%
\bibliographystyle{splncs04} 
\bibliography{ref}
\end{document}